\pdfoutput=1

\documentclass[11pt]{article}

\usepackage[final]{acl}

\usepackage{times}
\usepackage{latexsym}
\usepackage{kotex}
\usepackage{xcolor}
\usepackage{colortbl}

\usepackage{amsmath}
\usepackage{hyperref}
\usepackage{booktabs}
\usepackage{multirow}

\usepackage[T1]{fontenc}

\usepackage[utf8]{inputenc}

\usepackage{microtype}

\usepackage{inconsolata}

\usepackage{graphicx}

\title{MMRefine: Unveiling the Obstacles to Robust Refinement\\ in Multimodal Large Language Models}

\author{Gio Paik\thanks{Most work was done during the internship at NAVER Cloud AI.} \\
  Theta One, Inc. \\
  \texttt{giopaik@thetaone.co} \\\And
  Geewook Kim \\
  NAVER Cloud AI\\
  KAIST AI\\
  \texttt{gwkim.rsrch@gmail.com} \\\And
  Jinbae Im \thanks{Corresponding author}\\
  NAVER Cloud AI\\
  \texttt{jinbae.im@navercorp.com} \\}

\begin{document}
\maketitle
\begin{abstract}
This paper introduces MMRefine, a MultiModal Refinement benchmark designed to evaluate the error refinement capabilities of Multimodal Large Language Models (MLLMs).
As the emphasis shifts toward enhancing reasoning during inference, MMRefine provides a framework that evaluates MLLMs' abilities to detect and correct errors across six distinct scenarios beyond just comparing final accuracy before and after refinement.
Furthermore, the benchmark analyzes the refinement performance by categorizing errors into six error types.
Experiments with various open and closed MLLMs reveal bottlenecks and factors impeding refinement performance, highlighting areas for improvement in effective reasoning enhancement. Our code and dataset are publicly available at \url{https://github.com/naver-ai/MMRefine}.
\end{abstract}

\section{Introduction}

Recent advances have endowed MLLMs with remarkable capabilities, enabling them to tackle complex challenges such as mathematical reasoning and multimodal understanding~\citep{llama, llavanext, internvl25}. 
 
Instead of concentrating solely on scaling model parameters during training, current research aims to strengthen inference-time reasoning.
Techniques such as \textit{Self-Refinement}, where models iteratively improve their outputs~\citep{madaan2024self, li2024confidence, kumar2025training, huang2024query, zhang2024small}, and engaging multiple models or agents in debate to achieve consensus~\citep{liang2023encouraging, talebirad2023multi, chen2024magicore} have gained traction.

These methodologies heavily rely on the ability of MLLMs to evaluate and refine their responses.
If such capability is not sufficiently secured, refinement might unintentionally impair performance, causing incorrect corrections and unnecessarily prolonged response times~\citep{huang2023large}.
Therefore, it is essential to investigate whether MLLMs can accurately identify and correct errors in their reasoning processes.

However, previous studies primarily compare the accuracy of the final answer before and after applying refinement~\cite{huang2023large, li2024confidence} without sufficient analysis of MLLMs' refinement capabilities.
Although~\citet{yan2024errorradar} analyze specific abilities such as error localization and classification, their scope is limited to error detection capability.

\begin{figure}[t!]
  \includegraphics[width=\linewidth]{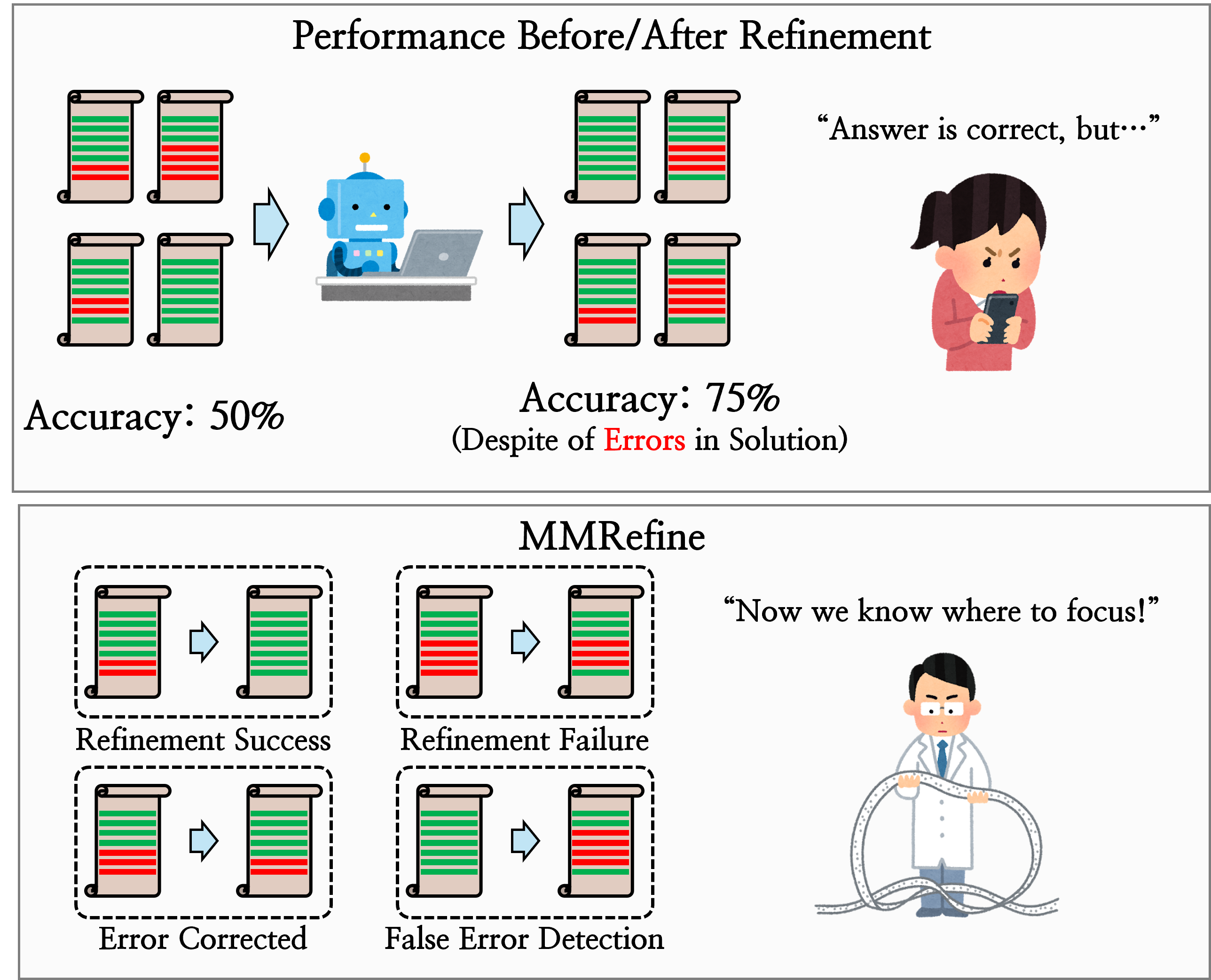}
  \caption{\textbf{Motivation.} Error refinement fails for a variety of scenarios and error types. Systematic evaluation is vital for providing accurate feedback and enhancing performance.}
  \label{fig:intro}
\end{figure}

\begin{figure}[t!]
  \includegraphics[width=\linewidth]{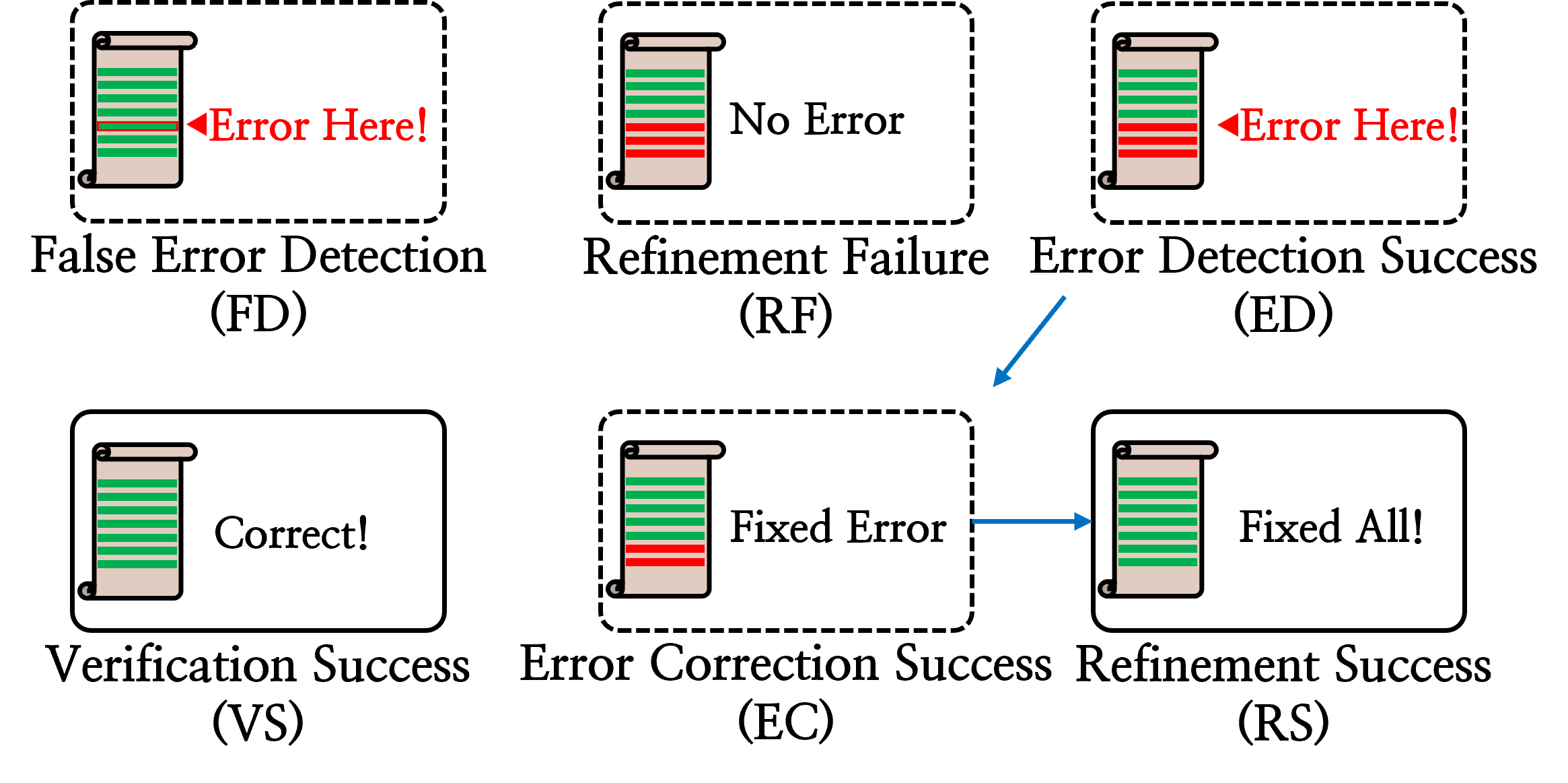}
  \caption{\textbf{Evaluation Protocol.} We define six scenarios to evaluate MLLM refinement capabilities.}
  \label{fig:eval_types}
\end{figure}

In this paper, we propose a new benchmark, \textbf{M}ulti\textbf{M}odal \textbf{Refine}ment (\textbf{MMRefine}), to evaluate whether MLLMs can detect errors in their initial solutions and improve them. 
Unlike previous studies that compare accuracy before and after refinement, MMRefine assesses refinement outcomes beyond mere final accuracy by categorizing them into six scenarios that can occur during the refinement process: False Error Detection and Verification Success for correct solutions, and Refinement Failure, Error Detection Success, Error Correction Success, and Refinement Success for incorrect solutions, as shown in~\autoref{fig:intro} and~\ref{fig:eval_types}.
Our approach enables the identification of refinement bottlenecks and offers a nuanced understanding of MLLMs' refinement capabilities.

Through extensive experiments, we validate that MMRefine is effective for evaluating and analyzing the refinement capability of MLLMs.
We evaluate $17$ MLLMs' refinement capability and examine which stages in the refinement process become bottlenecks for them.
By comparing these scores with the actual self-reflection results on other benchmarks, we demonstrate the potential of the MMRefine as a reliable benchmark for refinement ability.
Furthermore, we categorize the errors in MMRefine into six types and provide an analysis of the refinement performance for each error type. 
The analysis shows that MLLMs of various sizes and architectures exhibit varying strengths and weaknesses in correcting different types of errors.

Our study provides two main contributions.
First, we introduce MMRefine, a MultiModal Refinement benchmark designed to systematically evaluate the refinement capabilities of MLLMs across the entire refinement process.
Second, through comprehensive experiments, we evaluate the performance of each refinement process in MLLMs and analyze the error types to which they are vulnerable.

\begin{figure}[t!]
    \includegraphics[width=\linewidth]{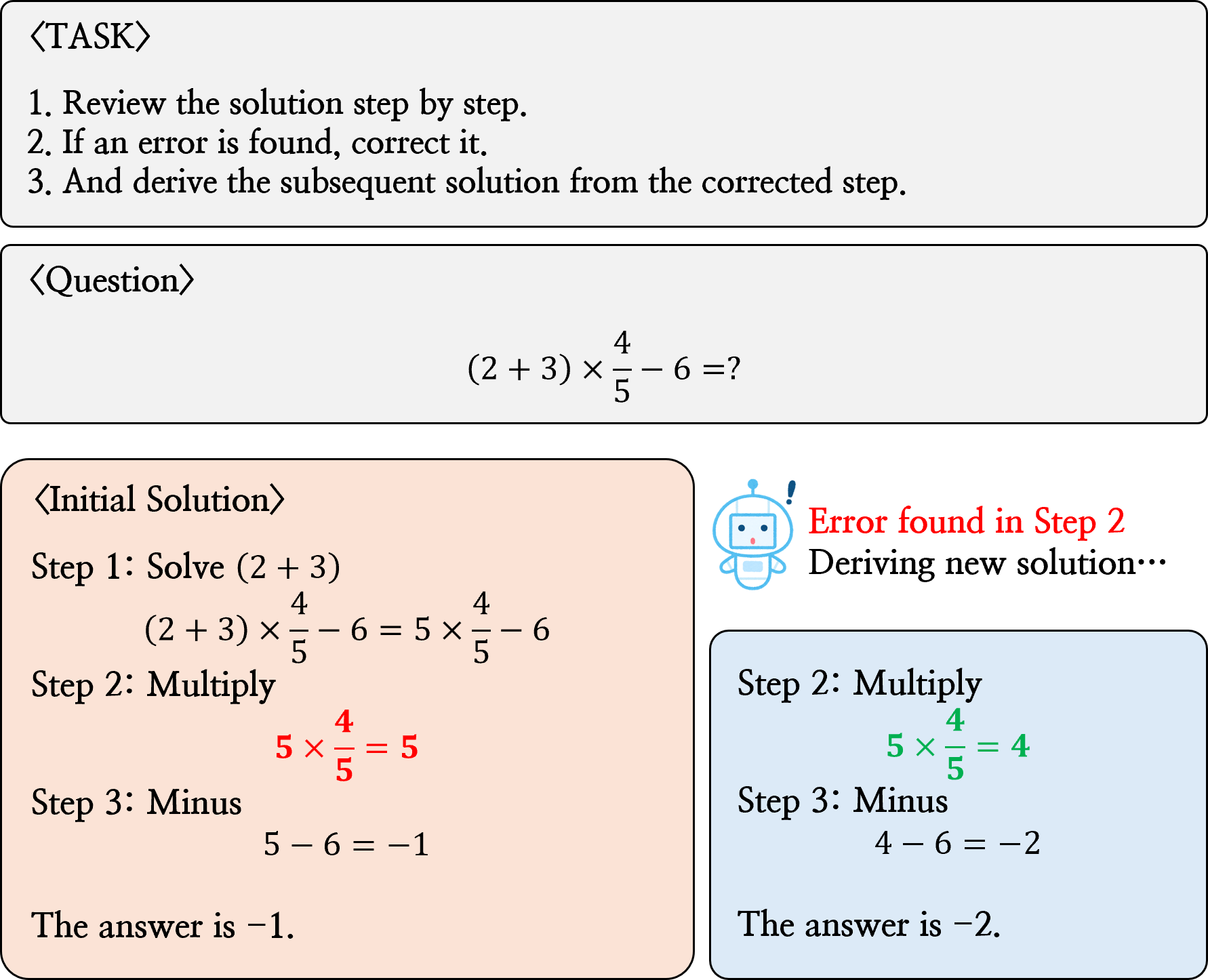}
    \caption{\textbf{Illustration of the Refinement Process.} The model identifies an \textcolor[rgb]{1.0,0.0,0.0}{\textbf{error}} in the initial solution and \textcolor[rgb]{0.0,0.5,0.0}{\textbf{corrects}} it, then proceeds to derive a revised solution from the point of correction.}
    \label{fig:toy_example}
\end{figure}

\section{MMRefine}

\subsection{Overview of MMRefine} 

We propose MMRefine, a novel benchmark that is meticulously designed to evaluate the refinement capability of MLLMs.
To effectively evaluate the models' ability to refine their responses, we focus on mathematical problems that are sufficiently challenging to require refinement, often leading to longer reasoning paths, and allow for a clear and logical determination of correctness. 
Given an initial solution to a problem, we prompt the model to generate an improved solution as shown in~\autoref{fig:toy_example}.
Unlike previous studies~\cite{huang2023large} that solely compared performance before and after refinement, we conduct a more granular evaluation by categorizing the refinement outcomes into six scenarios as depicted in~\autoref{fig:eval_types}.

More specifically, we instruct MLLMs to review the solution step-by-step, identify and correct any detected errors, and regenerate the solution from the corrected point.
Then, we categorize the refinement outcome into one of six scenarios by employing \textsc{GPT-4o} as a judge~\citep{llmjudge} and considering the reference feedback. To ensure the reliability of LLM-based evaluations, we perform human verification and \textsc{OpenAI o1} verification as described in~\autoref{sec:human_verification}.
If the model incorrectly identifies an error in a correct solution, it is classified as \textbf{False Error Detection (FD)}; otherwise, it is categorized as \textbf{Verification Success (VS)}.
If the model fails to detect an error in an incorrect solution, it is classified as \textbf{Refinement Failure (RF)}; otherwise, it is categorized as \textbf{Error Detection Success (ED)}.
Note that since errors propagate to subsequent steps, we focus on the first error.
Among the ED cases, if the error is accurately corrected, it is classified as \textbf{Error Correction Success (EC)}; furthermore, if the subsequent solution is generated flawlessly, it is categorized as \textbf{Refinement Success (RS)}.
For details on the refinement and evaluation process, please refer to~\autoref{sec:protocol_detail}.

\begin{table*}[t!]
\resizebox{\textwidth}{!}{%
\begin{tabular}{l|cccc|cc|cc}
\toprule
 & \textbf{\begin{tabular}[c]{@{}c@{}}Refinement \\ Failure (RF)\end{tabular}} & \textbf{\begin{tabular}[c]{@{}c@{}}Error Detection\\ Success (ED)\end{tabular}} & \textbf{\begin{tabular}[c]{@{}c@{}}Error Correction\\ Success (EC)\end{tabular}} & \textbf{\begin{tabular}[c]{@{}c@{}}Refinement\\ Success (RS)\end{tabular}} & \textbf{\begin{tabular}[c]{@{}c@{}}False Error\\ Detection (FD)\end{tabular}} & \textbf{\begin{tabular}[c]{@{}c@{}}Verification\\ Success (VS)\end{tabular}} & \textbf{RefScore} & \textbf{mRecall} \\ 
\midrule
\multicolumn{9}{c}{\textit{Closed-Source MLLMs}} \\
\midrule
\textsc{GPT-4o} \citep{gpt4o}             & 15.57 & 84.43 & 43.15 & 29.27 & \colorbox[HTML]{BCD4E6}{6.74} & \colorbox[HTML]{BCD4E6}{93.26} & 22.53 & \colorbox[HTML]{BCD4E6}{88.84} \\
\textsc{Gemini-1.5-Pro} \citep{google2024gemini15}    & \colorbox[HTML]{BCD4E6}{3.75} & \colorbox[HTML]{BCD4E6}{96.25} & \colorbox[HTML]{BCD4E6}{64.54} & \colorbox[HTML]{BCD4E6}{45.22} & 22.10 & 77.90 & \colorbox[HTML]{BCD4E6}{23.12} & 87.08 \\
\textsc{Claude-3.5-Sonnet} \citep{claude} & 27.95 & 72.05 & 32.65 & 18.95 & \colorbox[HTML]{BCD4E6}{6.74} & \colorbox[HTML]{BCD4E6}{93.26} & 12.21 & 82.65 \\ 
\midrule
\multicolumn{9}{c}{\textit{Open-Source MLLMs}} \\
\midrule
\textsc{LLaVA-OneVision-0.5B} \citep{llavaov}     & 36.40 & 63.60 & 2.06 & 2.06 & 75.66 & 24.34 & -73.59 & 43.97 \\
\textsc{InternVL2.5-1B} \citep{internvl25}        & 41.09 & 58.91 & 3.75 & 1.88 & 19.85 & 80.15 & -17.97 & 69.53 \\
\textsc{Qwen2-VL-2B} \citep{qwen2vl}              & 51.59 & 48.41 & 3.19 & 2.44 & 19.10 & 80.90 & -16.66 & 64.65 \\
\textsc{InternVL2.5-4B} \citep{internvl25}        & 45.22 & 54.78 & 6.00 & 4.13 & \colorbox[HTML]{ACE1AF}{0.75} & \colorbox[HTML]{ACE1AF}{99.25} & 3.38 & 77.02 \\
\textsc{LLaVA-NeXT-7B} \citep{llavanext}          & 42.40 & 57.60 & 5.44 & 4.13 & 4.49  & 95.51 & -0.37 & 76.55 \\
\textsc{LLaVA-OneVision-7B} \citep{llavaov}       & 42.59 & 57.41 & 5.44 & 4.50 & 1.87  & 98.13 & 2.63 & 77.77 \\
\textsc{Qwen2-VL-7B} \citep{qwen2vl}              & 19.70 & 80.30 &22.51 &\colorbox[HTML]{ACE1AF}{21.39} & 32.21 & 67.79 & -10.82 & 74.05 \\
\textsc{InternVL2.5-8B} \citep{internvl25}        & 25.14 & 74.86 &11.44 & 5.82 & 10.49 & 89.51 & -4.67 & 82.19 \\
\textsc{Llama-3.2-Vision-11B} \citep{llama}       & 22.14 & 77.86 &16.14 &10.51 & 32.96 & 67.04 & -22.45 & 72.45 \\
\textsc{Qwen2-VL-72B} \citep{qwen2vl}             & 20.26 & 79.74 &22.89 &13.70 & 20.60 & 79.40 & -6.90 & 79.57 \\
\textsc{LLaVA-NeXT-72B} \citep{llavanext}         & 22.14 & 77.86 &17.64 & 8.44 & 21.35 & 78.65 & -12.91 & 78.26 \\
\textsc{LLaVA-OneVision-72B} \citep{llavaov}      & 31.14 & 68.86 &21.76 & 11.07 & 4.87 & 95.13 & \colorbox[HTML]{ACE1AF}{6.20} & 81.99 \\ 
\textsc{InternVL2.5-78B} \citep{internvl25}       & \colorbox[HTML]{ACE1AF}{15.57} & \colorbox[HTML]{ACE1AF}{84.43} & \colorbox[HTML]{ACE1AF}{32.65} & 20.26 & 17.98 & 82.02 & 2.29 & \colorbox[HTML]{ACE1AF}{83.23} \\
\textsc{Llama-3.2-Vision-90B} \citep{llama}       & 16.89 & 83.11 & 28.33 & 16.51 & 17.23 & 82.77 & -0.72 & 82.94 \\
\bottomrule
\end{tabular}%
}
\caption{
    \textbf{MMRefine Benchmark Results.} The table shows performance metrics for closed-source and open-source MLLMs, with top scores highlighted in \colorbox[HTML]{BCD4E6}{blue} (closed-source) and \colorbox[HTML]{ACE1AF}{green} (open-source). Lower values are better for RF and FD, while higher values are better otherwise.
  }
\label{tab:bench_results_overall}
\end{table*}

\begin{table}[t!]
  \centering
\resizebox{0.48\textwidth}{!}{%
\begin{tabular}{l|cc|cc}
\toprule
\multirow{2}{*}{} & \multicolumn{2}{c|}{\textbf{Inference Time (s)}} & \multirow{2}{*}{\textbf{RefScore}} &\textbf{Refinement} \\
 & CoT & Refinement & &  \textbf{Efficiency}\\
\midrule
\textsc{GPT-4o}                  & 11.22 & 7.77 & 22.5 & 0.33\\
\textsc{Gemini-1.5-Pro}          & 7.67 & 7.78 & 23.1 & 0.23 \\
\textsc{Claude-3.5-Sonnet}       & 6.06 & 4.82 & 12.2 & 0.15 \\
\textsc{Llama-3.2-Vision-11B}    & 35.64 & 28.76 & -22.5 & - \\
\bottomrule
\end{tabular}%
}
\caption{\textbf{MMRefine Refinement Efficiency.} We calculate refinement efficiency by dividing RefScore by the percentage of refinement inference time relative to the initial CoT inference time.}
\label{tab:exp_efficiency}
\end{table}

\begin{table*}[t!]
  \centering
\resizebox{\textwidth}{!}{%
\begin{tabular}{l|cc|cc|cc|cc|cc}
    \toprule
    \multicolumn{1}{c|}{\multirow{2}{*}[-.3em]{}} & \multicolumn{2}{c|}{\textbf{MMRefine}} & \multicolumn{2}{c|}{\textbf{MMRefine$_\text{Text-only}$}} & \multicolumn{2}{c|}{\textbf{MATH-500}} & \multicolumn{2}{c|}{\textbf{MMRefine$_\text{Visual}$}} & \multicolumn{2}{c}{\textbf{MathVista}} \\
    \textbf{} & RefScore & mRecall & RefScore & mRecall & CoT & Self-Reflection($\Delta$) & RefScore & mRecall & CoT & Self-Reflection($\Delta$) \\ 
    \midrule
    \textsc{GPT-4o}                  & 22.5 & 88.8 & 33.8 & 93.8 & 73.4 & 75.2 (+1.8) & 12.9 & 84.5 & 60.5 & 61.2 (+0.7)\\
    \textsc{Gemini-1.5-Pro}          & 23.1 & 87.1 & 45.1 & 92.5 & 79.8 & 80.6 (+0.8) & -8.8 & 74.5 & 71.6 & 70.6 (-1.0)\\
    \textsc{Claude-3.5-Sonnet}       & 12.2 & 82.7 & 21.3 & 88.3 & 61.2 & 62.2 (+1.0) & 3.9  & 77.8 & 63.0 & 63.2 (+0.2)\\
    \textsc{Llama-3.2-Vision-11B}    & -22.5& 72.5 & -16.8& 80.7 & 37.4 & 37.4 (0.0) & -13.7 & 73.6 & 48.4 & 47.3 (-1.1)\\
    \bottomrule
    \end{tabular}%
}
  \caption{\textbf{Comparison of MMRefine Scores and the Self-Reflection Results in MATH-500 and MathVista.} To conduct an in-depth analysis of the results in text-only and visual math problems, we report the results for the two subsets of MMRefine: $\text{MMRefine}_\text{Text-only}$ consisting of MathOdyssey problems and $\text{MMRefine}_\text{Visual}$ consisting of MathVision problems. Refer to~\autoref{sec:exp_detail_otherbench} for details.}
  \label{fig:performance_comparison}
\label{tab:self-reflection}
\end{table*}

\begin{table*}[t!]
\resizebox{\textwidth}{!}{%
\begin{tabular}{l|*{4}{>{\centering\arraybackslash}p{2.5cm}}@{\hspace{4pt}}|@{\hspace{4pt}}*{2}{>{\centering\arraybackslash}p{2.5cm}}}
\toprule
 & \textbf{Problem} & \textbf{Logical} & \multirow{2}{*}{\textbf{Calculation}} & \multirow{2}{*}{\textbf{Equation}} & \textbf{Visual} & \textbf{Spatial} \\
  & \textbf{Understanding} & \textbf{Reasoning} & & & \textbf{Perception} & \textbf{Reasoning} \\
\midrule
\multicolumn{7}{c}{\textit{Closed-Source MLLMs}} \\
\midrule
\textsc{GPT-4o} \citep{gpt4o}                       & \colorbox[HTML]{BCD4E6}{36.7} & 29.4 & 32.8 & 34.7 & 26.3 & \colorbox[HTML]{FBCCE7}{11.5} \\
\textsc{Gemini-1.5-Pro} \citep{google2024gemini15}  & 36.7 & 48.6 & \colorbox[HTML]{BCD4E6}{67.2} & 61.2 & 35.0 & \colorbox[HTML]{FBCCE7}{23.1} \\
\textsc{Claude-3.5-Sonnet} \citep{claude}           & 25.0 & 22.0 & 18.0 & \colorbox[HTML]{BCD4E6}{28.6} & 13.8 & \colorbox[HTML]{FBCCE7}{0.0}  \\
\midrule
\multicolumn{7}{c}{\textit{Open-Source MLLMs}} \\
\midrule
\textsc{LLaVA-OneVision-0.5B} \citep{llavaov}       & 3.3  & 0.6  & \colorbox[HTML]{BCD4E6}{4.9}  & \colorbox[HTML]{FBCCE7}{0.0}  & 2.5  & 3.8  \\
\textsc{InternVL2.5-1B} \citep{internvl25}          & 1.7  & 1.7  & 1.6  & \colorbox[HTML]{FBCCE7}{0.0}  & 1.9  & \colorbox[HTML]{BCD4E6}{7.7}  \\
\textsc{Qwen2-VL-2B} \citep{qwen2vl}                & \colorbox[HTML]{FBCCE7}{0.0}  & 2.8  & 3.3  & \colorbox[HTML]{FBCCE7}{0.0}  & 2.5  & \colorbox[HTML]{BCD4E6}{7.7}  \\
\textsc{InternVL2.5-4B} \citep{internvl25}          & 3.3  & 4.0  & \colorbox[HTML]{BCD4E6}{6.6}  & 6.1  & \colorbox[HTML]{FBCCE7}{3.1}  & 3.8  \\
\textsc{LLaVA-NeXT-7B} \citep{llavanext}            & 5.0  & \colorbox[HTML]{FBCCE7}{1.7}  & 3.3  & 2.0  & 3.8  & \colorbox[HTML]{BCD4E6}{26.9} \\
\textsc{LLaVA-OneVision-7B} \citep{llavaov}         & 3.3  & \colorbox[HTML]{FBCCE7}{2.8}  & 4.9  & 4.1  & 4.4  & \colorbox[HTML]{BCD4E6}{19.2} \\
\textsc{Qwen2-VL-7B} \citep{qwen2vl}                & 11.7 & 19.8 & 26.2 & \colorbox[HTML]{FBCCE7}{10.2} & 26.3 & \colorbox[HTML]{BCD4E6}{34.6} \\
\textsc{InternVL2.5-8B} \citep{internvl25}          & 5.0  & 4.5  & \colorbox[HTML]{BCD4E6}{8.2}  & \colorbox[HTML]{BCD4E6}{8.2}  & 6.3  & \colorbox[HTML]{FBCCE7}{0.0}  \\
\textsc{Llama-3.2-Vision-11B} \citep{llama}         & 6.7  & 15.8 & \colorbox[HTML]{BCD4E6}{18.0} & 16.3 & \colorbox[HTML]{FBCCE7}{2.5}  & 3.8  \\
\textsc{Qwen2-VL-72B} \citep{qwen2vl}               & \colorbox[HTML]{FBCCE7}{8.3}  & 11.9 & \colorbox[HTML]{BCD4E6}{21.3} & 12.2 & 15.6 & 11.5 \\
\textsc{LLaVA-NeXT-72B} \citep{llavanext}           & 8.3  & \colorbox[HTML]{FBCCE7}{7.3}  & 9.8  & \colorbox[HTML]{BCD4E6}{12.2} & 8.1  & 7.7  \\
\textsc{LLaVA-OneVision-72B} \citep{llavaov}        & \colorbox[HTML]{BCD4E6}{15.0} & 11.9 & 14.8 & 8.2  & 8.8  & \colorbox[HTML]{FBCCE7}{7.7}  \\
\textsc{InternVL2.5-78B} \citep{internvl25}         & 16.7 & 20.9 & \colorbox[HTML]{BCD4E6}{26.2} & 16.3 & 21.3 & \colorbox[HTML]{FBCCE7}{11.5} \\
Llama-3.2-Vision-90B \citep{llama}                  & \colorbox[HTML]{FBCCE7}{15.0} & 16.4 & \colorbox[HTML]{BCD4E6}{19.7} & 18.4 & 15.6 & 15.4 \\
\bottomrule
\end{tabular}%
}
\caption{
    \textbf{Comparison of RefScore by First Error Type.} \colorbox[HTML]{BCD4E6}{Maximum} and \colorbox[HTML]{FBCCE7}{minimum} values are highlighted.}
\label{tab:error_type_ablation}
\end{table*}

\subsection{Evaluation Metrics}

To analyze the bottleneck stages during the refinement process, we calculate the proportions for each result scenario.
Since the ratio of incorrect solutions to correct solutions differs, we separately measure the ratios of FD and VS among the correct solutions and the ratios of RF, ED, EC, and RS among the incorrect solutions.

For straightforward comparison and evaluation of refinement capabilities, we introduce \textbf{RefScore}, a metric concentrating on the overall refinement performance of MLLMs, and \textbf{mRecall}, a metric emphasizing error detection performance.
The RefScore is defined as:
$$ \text{RefScore} = \text{RS} - \text{FD} $$
where RS and FD represent the proportions of corrected and uncorrected solutions, respectively.
Meanwhile, mRecall is defined as: $$ \text{mRecall} = (\text{ED} + \text{VS}) / 2.$$
This measures the model's ability to both detect actual errors and verify correct solutions accurately.

\subsection{Dataset Construction}
\label{sec:dataset_construction}

We construct MMRefine by carefully curating both text-only and visual math problems.
We sample $100$ text-only problems from MathOdyssey~\citep{mathodyssey} and $100$ visual problems from MathVision~\citep{mathvision} covering various subjects and levels of difficulty as described in~\autoref{sec:dataset_details}.

To provide a variety of initial solutions, we generate total $800$ initial solutions using four MLLMs: \textsc{GPT-4o}~\citep{gpt4o}, \textsc{Gemini-1.5-Pro}~\citep{google2024gemini15,gemini}, \textsc{Claude-3.5-Sonnet}~\citep{claude}, and \textsc{Llama-3.2-Vision-11B}~\citep{llama}.
Note that MMRefine evaluates refinement processes under realistic conditions. 
Unlike previous studies that generated initial solutions by adding errors to correct solutions~\citep{nath2025can, li2024evaluating} or imposing constraints such as limiting the chain-of-thought steps on LLMs~\citep{wu2024visco} to evaluate refinement capabilities, we employ solutions generated without any constraints.  

For reliable evaluation, reference feedbacks are generated by \textsc{OpenAI o1}~\citep{openaio1} using the original human-annotated solutions, and we validate them through the revision process.
We test whether three MLLMs (\textsc{GPT-4o}, \textsc{Gemini-1.5-Pro}, and \textsc{Claude-3.5-Sonnet}) can revise incorrect initial solutions when reference feedbacks are provided.
We retain only the feedback where refinement success is confirmed across all models and ensure validity by either regenerating or manually correcting the flawed feedback.

To conduct a detailed analysis of MLLMs' refinement capabilities based on error types, we manually categorize the first errors in the initial solutions into six types as depicted in~\autoref{sec:human_annotation_detail}.
Detailed explanations of six error types are detailed in~\autoref{sec:error_types}. 

\section{Experiments}
\subsection{Overall Performance}
We evaluate $17$ MLLMs, including $3$ closed-source models and various open-source models from $5$ MLLM families of differing sizes, as shown in~\autoref{tab:bench_results_overall}. In terms of mRecall, closed-source models demonstrate superior performance compared to all open-source models below 11B. Only a few large-scale open-source models, namely \textsc{InternVL2.5-78B} and \textsc{Llama-3.2-Vision-90B} manage to surpass \textsc{Claude-3.5-Sonnet}.
Additionally, closed-source models correct errors in over $32$\% of initial solutions, while most open-source models correct less than $20$\%. For RS, which reflects perfect response improvement, only two open-source models, \textsc{InternVL2.5-78B} and \textsc{Qwen2-VL-7B}, exceed the performance of the lowest closed-source records. Despite this, their RefScores remain lower than closed-source models due to high FD. 
These findings suggest that the current error correction proficiency of most open-source models remains inadequate for effective refinement, even in larger models exceeding 70B parameters. However, notable exceptions including \textsc{Qwen2-VL-7B}, which achieves a higher RS score than \textsc{Claude-3.5-Sonnet}, and \textsc{InternVL2.5-8B}, which records a high mRecall score of $82.19$, indicate refinement potentials even in mid-scale models.

To ascertain which MLLMs offer reasonable refinement performance relative to the increased computational cost caused by refinement, we also measure the refinement efficiency, as shown in~\autoref{tab:exp_efficiency}.
Adding the refinement step increases inference time by $60$-$100$\% compared to the initial CoT inference. Notably, refinement efficiency differs significantly between models. 
Although \textsc{Gemini-1.5-Pro} achieves a higher RefScore compared to \textsc{GPT-4o}, the refinement efficiency of \textsc{GPT-4o} is substantially higher.
In practice, adopting refinement necessitates balancing the increase in inference time with the anticipated performance gain.

\subsection{Correlation with Self-Reflection}
\label{sec:otherbench}
We also analyze the correlation between MMRefine scores and score changes after self-reflection in other math benchmarks: MATH-500~\cite{mathbench} and MathVista~\cite{mathvista}, as shown in~\autoref{tab:self-reflection}.
The results show that RefScore and mRecall are correlated with the models' refinement capability.
With the exception of \textsc{Gemini-1.5-Pro}, RefScores in text-only and visual math problems are directly correlated with the score changes in MATH-500 and MathVista (correlation coefficient $0.82$).
\textsc{Gemini-1.5-Pro}, particularly for visual problems, exhibits a relatively low mRecall, which appears to have led to a decrease in scores after self-reflection in MathVista.
From the results, we demonstrate that the MMRefine scores are valuable indicators of the refinement capability.

\subsection{Error Type Analysis}

To understand what types of errors impede effective refinement, we analyze the RefScore by six distinct error types, as detailed in~\autoref{tab:error_type_ablation}. 
While different MLLMs exhibit varying strengths and weaknesses in refining specific error types, our findings indicate that larger models with higher capacities tend to perform significantly better at correcting four text-related errors than two image-related ones. In contrast, models smaller than 7B often demonstrate superior handling of image-related errors. 
For instance, \textsc{LLaVA-Next-7B} and \textsc{Qwen2-VL-7B} perform better than even closed-source models in correcting spatial reasoning errors.
While this discrepancy may be partially attributed to differences in LLM and vision encoder sizes, it could also be influenced by the curriculum through which MLLMs acquire their capabilities.

\begin{figure}[t!]
    \includegraphics[width=\linewidth]{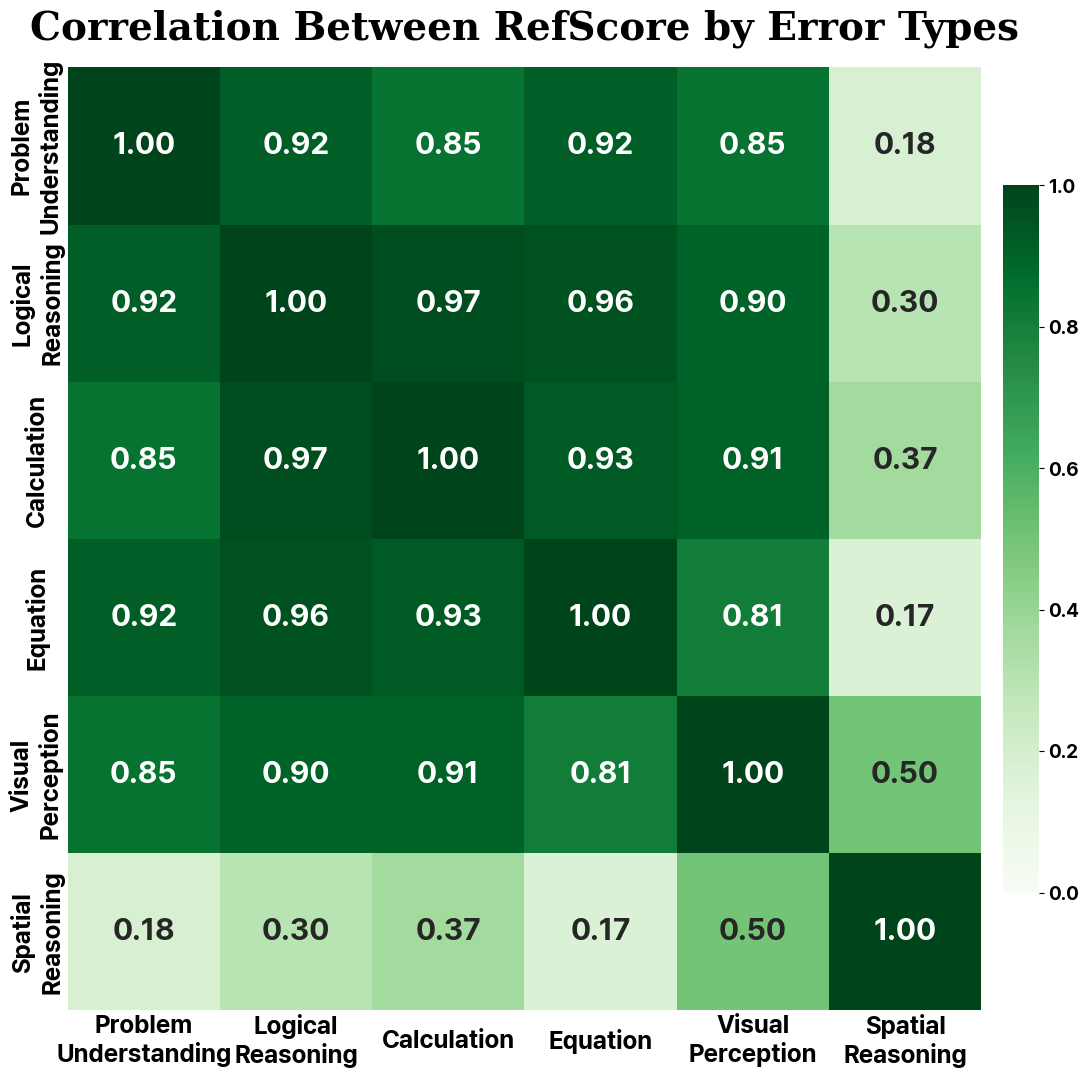}
    \caption{\textbf{Correlation Between Refscore by Error Types.} We calculate the correlation coefficients of RefScore for each error type across $17$ models.}
\label{fig:corr_error_type}
\end{figure}

We also compare the correlation between RefScores for each error type. As shown in~\autoref{fig:corr_error_type}, the performance on most error types is highly correlated with that on other error types, 
whereas spatial reasoning error type shows low correlations overall.
This observation suggests that there may be alternative approaches to enhancing refinement performance for specific error types, such as spatial reasoning errors, beyond merely scaling up the refinement capability of MLLMs.

\subsection{LLM-based Evaluation}
\label{sec:human_verification}
The nature of solving math problems allows for diverse approaches and infinitely varied errors and correction methods within the reasoning process. Because the human evaluation of such complex reasoning is not only highly demanding but can even be inaccurate, automated methods such as LLM-as-a-Judge~\cite{llmjudge} have been proposed to evaluate MLLMs' reasoning processes. In MMRefine, we use \textsc{GPT-4o} as a judge. To ensure the reliability of it, we compare \textsc{GPT-4o}'s judgments with those from human evaluations and \textsc{OpenAI o1}. \textsc{GPT-4o} achieves $72$\% agreement with human judgments and $73$\% agreement with the \textsc{OpenAI o1} judgments. While the alignments are not perfect, the practical advantage of LLM-based evaluation becomes clear when considering its efficiency and scalability. Whereas a human evaluation conducted by an expert with university-level mathematical knowledge takes over $8$ hours, \textsc{GPT-4o} provides reliable judgments in much shorter time and effort.

\section{Conclusion}
This paper introduces MMRefine, which evaluates the refinement capabilities of MLLMs through an analysis of their outcomes across six distinct scenarios and six error types.
Our comprehensive assessment of $17$ MLLMs reveals that larger models tend to refine textual errors better, whereas smaller models are more effective with visual errors. For spatial reasoning errors, only specific models exhibit a certain level of refinement capability.
These insights into intrinsic refinement capabilities can enhance MLLMs' reasoning abilities and provide guidance for addressing their weaknesses.

\section*{Limitations}
In this paper, we generate evaluation data by solving problems collected from two math benchmark datasets using four closed-source and open-source MLLMs. While this approach allows for consistent evaluation of MLLM refinement capability, the resulting data inherently lacks the diversity of real-world use case scenarios and the breadth of initial solutions common in practice, such as solutions from other models, human-authored responses, and non-mathematical reasoning processes.
Furthermore, although various correct answers may exist when solving math problems, we adopt a single reference solution and conduct rigorous evaluations to facilitate LLM-based assessment. This bias can lead the model to be robust only in a few mathematical reasoning methods and overlook other important issues (e.g., fact verification, diverse and original problem-solving approaches). Although we inevitably conduct evaluations based on a limited reference solution to enhance the credibility of LLM-as-judge assessment, we aim to explore more flexible methods for evaluating and verifying the reasoning validity of MLLMs in future work.

\section*{Ethical Considerations}
We acknowledge that, due to practical considerations, the experimental results detailed in this paper are derived from single-run assessments.  
However, to uphold the reliability of our evaluation framework, we dedicate considerable effort to refine the prompts for the LLM-based judges carefully. Additionally, for the selected model, most prominently \textsc{Gemini-1.5-Pro}, we execute evaluations across three iterations and observe that the standard deviation of the resulting RefScore remains comfortably below $1$, thus suggesting a degree of score stability.

\bibliography{custom}

\clearpage

\appendix

\section{Refinement \& Evaluation Protocol Details}
\label{sec:protocol_detail}
In this section, we delve into the specifics of how refinement outcomes are generated and evaluated using the MMRefine benchmark. Initially, we provide the model with a math problem and an initial solution and instruct it to review the solution step-by-step using the prompt in~\autoref{fig:refinement_prompt}. 
If an error is identified during the review, the model stops reviewing and performs refinement starting from that step. At the end of refinement, the model provides the following outputs: the correctness of the initial solution, the explanation for this determination, and the final answer. 

When the initial solution refined by the model is indeed correct, we determine whether the refinement result is Verification Success (VS) or False Error Detection (FD) by parsing the correctness from the model's response, as shown in~\autoref{fig:correct_eval_prompt}. 
Conversely, if the initial solution is incorrect, we evaluate the model's refinement outcome by comparing it with the reference feedback through the prompt in~\autoref{fig:incorrect_eval_prompt}. Specifically, the model's success begins with Error Detection (ED), which is a prerequisite for subsequent scenarios, assessed based on the ``Error Detection'' rubric of the prompt. Following successful error detection, the model's Error Correction (EC) capability is evaluated using the ``Error Correction'' rubric. Finally, Refinement Success (RS), determined based on the ``Effectiveness and Correctness of the Feedback'' rubric, signifies that the model not only detects and corrects the error(s) but also derives a correct solution to the correct answer, encompassing both ED and EC.

\section{Datasets Details}
\label{sec:dataset_details}
We sample $200$ mathematical questions from the MathOdyssey~\citep{mathodyssey} and MathVision~\citep{mathvision} dataset as described in~\autoref{sec:dataset_construction}. The MathOdyssey dataset features mathematical questions from a wide range of subjects, encompassing difficulty levels from High School to University and Olympiad. 
Conversely, the MathVision dataset offers visual math problems across various domains, categorized by difficulty levels $1$ through $5$. We curate $100$ problems from each of these two datasets to construct our benchmark, as summarized in~\autoref{tab:data_statistics}. 

MathOdyssey dataset is distributed under the CC BY-SA 4.0 license, which permits its use as a test set. 
The license covers the dataset itself but not questions in the dataset. 
The MathVista dataset is available under the MIT License. 
MMRefine, derived from two benchmarks, is released under the CC BY-SA 4.0 license. 
This license covers the elements we create or label, while the copyright of the original questions remains with their respective authors.  
Similar to MathOdyssey, MMRefine is also restricted for testing purposes only, and its use as training data for models is prohibited.

\begin{table}[t!]
\resizebox{0.5\textwidth}{!}{%
\begin{tabular}{l|c|c}
\toprule[1.5px]
\multirow{17}{*}{MathOdyssey} & \textbf{Subject} & \textbf{\# of Questions} \\ \cmidrule{2-3}
& Algebra & 39 \\
& Precalculus & 12 \\
& Geometry & 11 \\
& Combinatorics & 10 \\
& Linear Algebra And Abstract Algebra & 7 \\
& Calculus And Analysis & 6 \\
& Probability & 5 \\
& Differential Equations & 4 \\
& Statistics & 4 \\
& Number Theory & 1 \\
& Calculus & 1 \\ 
 \cmidrule{2-3}
 & \textbf{Level} & \textbf{\# of Problems} \\ \cmidrule{2-3}
 & High School Math & 35 \\
 & High School Competition & 39 \\
 & College Math & 26 \\ \midrule[1.3pt]
\multirow{17}{*}{MathVision} & \textbf{Subject} & \textbf{\# of Questions} \\ \cmidrule{2-3}
& Metric Geometry & 48 \\
& Solid Geometry & 13 \\
& Combinatorial Geometry & 7 \\
& Algebra & 6 \\
& Transformation Geometry & 6 \\
& Descriptive Geometry & 6 \\
& Combinatorics & 5 \\
& Graph Theory & 3 \\
& Logic & 3 \\
& Arithmetic & 2 \\
& Counting & 1 \\
 \cmidrule{2-3}
 & \textbf{Level} & \textbf{\# of Problems} \\ \cmidrule{2-3}
 & Level 2 & 25 \\
 & Level 3 & 30 \\
 & Level 4 & 29 \\
 & Level 5 & 16 \\
\bottomrule[1.5pt]
\end{tabular}%
}
    \caption{
        \textbf{MMRefine Data Statistics.} MMRefine consists of problems that cover a wide range of subjects and levels of difficulty.
    }
\label{tab:data_statistics}
\end{table}

\begin{table*}[t!]
\resizebox{\textwidth}{!}{%
\begin{tabular}{l*{4}{>{\centering\arraybackslash}p{3cm}}}
\toprule[1.5pt]
 & \multicolumn{4}{c}{\textbf{Source of Initial Solution}} \\
 & \textbf{\textsc{GPT-4o}} & \textbf{\textsc{Gemini-1.5-Pro}} & \textbf{\textsc{Claude-3.5-Sonnet}} & \textbf{\textsc{Llama-3.2-Vision}} \\ \midrule
    \textsc{GPT-4o} & \cellcolor[HTML]{ACE1AF}20.97 & 14.82 & \cellcolor[HTML]{FBCCE7}16.90 & \cellcolor[HTML]{BCD4E6}33.41 \\
    \textsc{Gemini-1.5-Pro} & \cellcolor[HTML]{FBCCE7}18.25 & 11.86 & \cellcolor[HTML]{ACE1AF}21.79 & \cellcolor[HTML]{BCD4E6}38.06 \\
    \textsc{Claude-3.5-Sonnet} & \cellcolor[HTML]{BCD4E6}19.65 & \cellcolor[HTML]{FBCCE7}7.06 & 3.69 & \cellcolor[HTML]{ACE1AF}18.49 \\
    \textsc{Llama-3.2-Vision-11B} & \cellcolor[HTML]{BCD4E6}-15.83 & \cellcolor[HTML]{FBCCE7}-24.51 & -30.01 & \cellcolor[HTML]{ACE1AF}-19.50 \\ \bottomrule[1.5pt]
\end{tabular}%
}
\caption{
    \textbf{Performance Comparison of MLLMs Across Different Models Generating Initial Solutions.} In each row, the highest RefScore is highlighted in \colorbox[HTML]{BCD4E6}{blue}, the second highest in \colorbox[HTML]{ACE1AF}{green}, and the third highest in \colorbox[HTML]{FBCCE7}{pink}.
  }
\label{tab:solution_source_ablation}
\end{table*}

\section{Human Annotations Details} 
\label{sec:human_annotation_detail}
As detailed in~\autoref{sec:dataset_construction}, we manually annotate the first error type in each initial solution.
Annotators are tasked with labeling each initial solution, referencing the math problem, the model's generated solution, and the problem's original solution to determine the presence of errors and, if errors are found, to categorize the first error by its type.
The annotation is conducted by $14$ annotators, with $12$ holding a bachelor's degree and 2 holding a master's degree.

\section{Detailed Explanation of Error Types}
\label{sec:error_types}
\begin{figure*}[t!]
  \includegraphics[width=\textwidth]{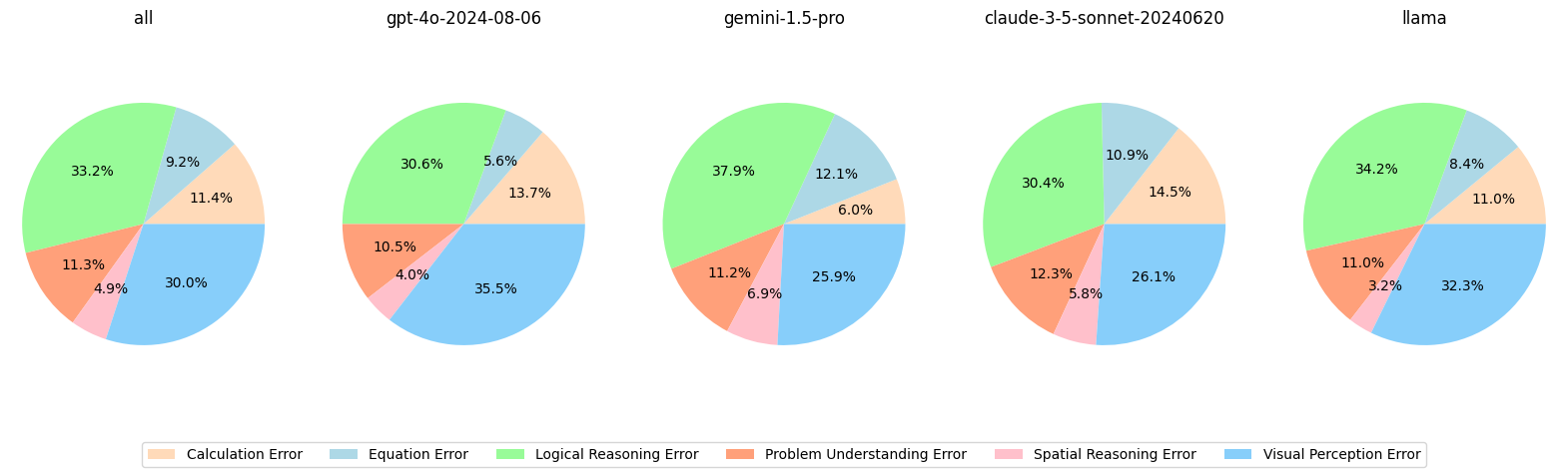}
  \caption{\textbf{Error Type Distribution of Initial Solutions by Model}}
  \label{fig:error_distribution}
\end{figure*}

\begin{figure*}[t!]
  \includegraphics[width=\textwidth]{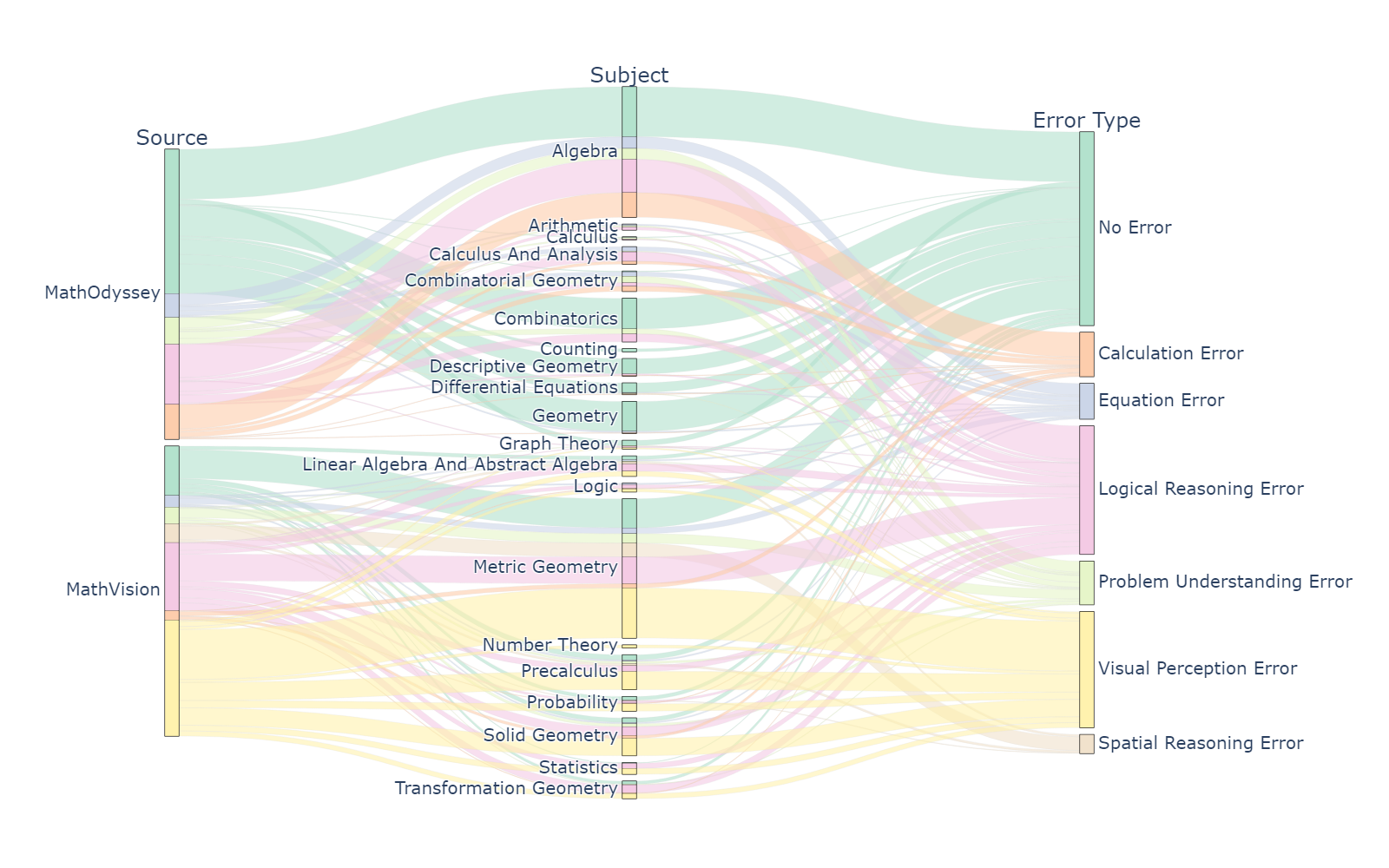}
  \caption{\textbf{Problem Source, Subject, and Error Type Distribution in MMRefine}}
  \label{fig:problem_distribution}
\end{figure*}

To enable a nuanced analysis of MLLMs' refinement capabilities across various situations, particularly concerning the nature of errors, we implement a categorization scheme encompassing six distinct error types.
\textbf{Problem understanding error} occurs when the model misinterprets the instructions or constraints explicitly stated in the problem description. 
\textbf{Logical reasoning error} denotes instances where the solution exhibits a flaw in the logical flow of argumentation, leading to an invalid conclusion. 
\textbf{Calculation error} refers to inaccuracies arising from numerical computation mistakes within the mathematical derivation. 
\textbf{Equation error} encompasses a range of mistakes related to algebraic manipulation, including, but not limited to, incorrect equation expansion or invalid variable substitution. 
\textbf{Visual perception error} is identified when the model fails to correctly interpret or recognize essential information conveyed through the problem's accompanying image.  
Lastly, \textbf{spatial reasoning error} is characterized by errors stemming from flawed spatial reasoning, such as incorrect assessments of geometric relationships or misinterpretations of spatial limitations. The error type distribution of MMRefine is presented in~\autoref{fig:error_distribution} and~\autoref{fig:problem_distribution}.

\section{Experimental Details for~\autoref{sec:otherbench}}
\label{sec:exp_detail_otherbench}
To explore the correlation of MMRefine with existing benchmarks, we conduct self-reflection experiments on MATH~\citep{mathbench} and MathVista~\citep{mathvista}, prominent benchmarks within the Large Language Model research community. For MATH, we perform evaluations using the $500$ test subset, as used in~\citep{lightman2024lets}. For MathVista, evaluations are conducted on the testmini set. We begin by evaluating the model's baseline performance using basic Chain-of-Thought (CoT) prompting~\citep{cot}. Subsequently, we prompt the model to refine its initial response through self-reflection, utilizing the prompt detailed in~\autoref{fig:self_reflection_prompt}. 

\section{RefScore Comparison by Solution Source}
\begin{figure*}[htbp]
  \includegraphics[width=\textwidth]{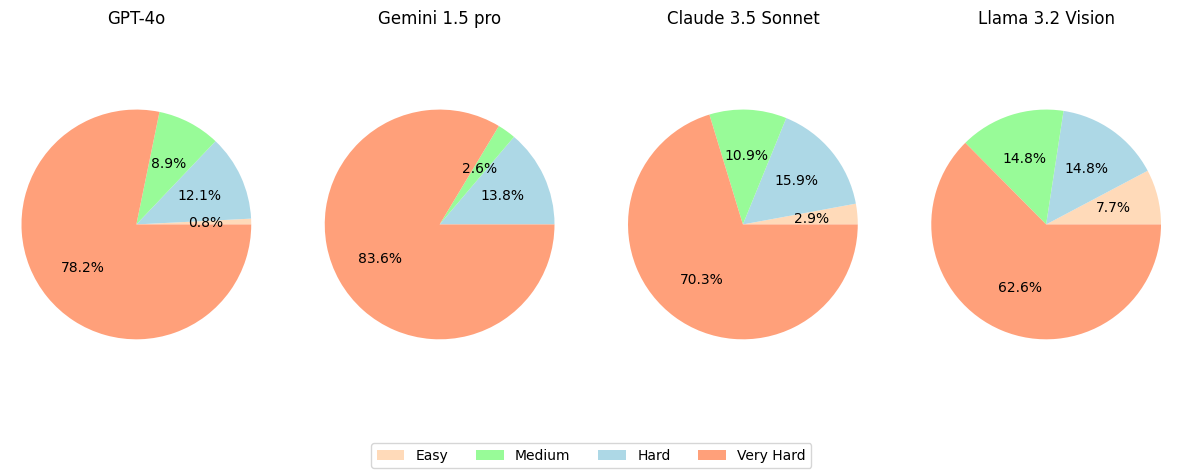}
  \caption{\textbf{Difficulty Distribution of Problems for Which Initial Solutions Are Incorrect by Model.}  Problem difficulty is determined by the number of MLLMs that correctly solve it. Specifically, if three out of the four models (\textsc{GPT-4o}, \textsc{Gemini-1.5-Pro}, \textsc{Claude-3.5-Sonnet}, and \textsc{Llama-3.2-Vision}) solve a problem correctly, the difficulty is categorized as `Easy'. If two models solve it, the difficulty is `Medium' and so on.}
  \label{fig:level_distribution}
\end{figure*}

We conduct experiments to investigate how refinement efficacy varies depending on the model that provides the initial solution. 
As shown in~\autoref{tab:solution_source_ablation}, all models achieve their best RefScore from initial solutions originating from \textsc{Llama-3.2-Vision} or \textsc{GPT-4o}. 
Interestingly, most models tend to successfully refine initial solutions generated by \textsc{Llama-3.2-Vision}.  
A plausible interpretation for this trend is that \textsc{Llama-3.2-Vision} tends to generate responses with errors skewed towards easier problem instances, thereby facilitating more effective refinement, as shown in~\autoref{fig:level_distribution}.

\begin{table*}[ht!]
  \centering
\resizebox{0.9\textwidth}{!}{%

\begin{tabular}{l|cccc|cc|cc}
\toprule
& RF ($\downarrow$) & ED ($\uparrow$) & EC ($\uparrow$) & RS ($\uparrow$) & VS ($\uparrow$) & FD ($\downarrow$) & RefScore & mRecall \\
\midrule
\textsc{GPT-4o}                     & 15.57 & 84.43 & 43.15 & 29.27 & 93.26 & 6.74 & 22.53 & 88.84 \\
\textsc{+VisualPRM-8B (N=4)}        & 20.83 & 79.17 & 40.53 & 22.33 & \textbf{96.63} & \textbf{3.37} & 18.96 & 87.90 \\
\textsc{Gemini-1.5-Pro}             & 3.75 & 96.25 & 64.54 & 45.22 & 77.90 & 22.10 & 23.12 & 87.08 \\
\textsc{+VisualPRM-8B (N=4)}        & 7.88 & 92.12 & 58.91 & 39.59 & \textbf{82.40} & \textbf{17.60} & 21.98 & \textbf{87.26} \\
\textsc{Claude-3.5-Sonnet}          & 27.95 & 72.05 & 32.65 & 18.95 & 93.26 & 6.74 & 12.21 & 82.65 \\ 
\textsc{+VisualPRM-8B (N=4)}        & 31.52 & 68.48 & \textbf{33.58} & 18.76 & \textbf{95.51} & \textbf{4.49} & \textbf{14.27} & 81.99 \\
\textsc{Llama-3.2-Vision-11B}       & 22.14 & 77.86 & 16.14 & 10.51 & 67.04 & 32.96 & -22.45 & 72.45 \\
\textsc{+VisualPRM-8B (N=4)}        & 25.52 & 74.48 & \textbf{21.95} & \textbf{13.32} & \textbf{80.15} & \textbf{19.85} & \textbf{-6.53} & \textbf{77.32} \\
\bottomrule
\end{tabular}%
}
  \caption{\textbf{MMRefine Performance Before and After Applying Best-of-N Selection with the VisualPRM.} Improved values are highlighted in \textbf{bold}.}
\label{tab:exp_prm_baseline}
\end{table*}

\section{Comparison with Process Reward Models}

We evaluate whether RefScore correlates with Process Reward Models (PRMs), which are used to assess MLLMs' reasoning processes and select better ones.
As shown in~\autoref{tab:exp_prm}, we calculate the correlation between RefScore and the directional changes in rewards of \textsc{VisualPRM-8B}~\cite{wang2025visualprm} before and after refinement.
Our experimental results show a moderate relationship between assessing improvements from refinements using PRM rewards and the RefScore. This finding can shed light on new directions for future research to analyze and enhance the performance of reward models in selecting better responses, particularly from the refinement perspective.

We also examine whether PRMs could also improve the refinement process, as shown in~\autoref{tab:exp_prm_baseline}.
After applying the best-of-N selection with the \textsc{VisualPRM-8B}, we observe a trade-off where Error Detection (ED) decreases while Verification Success (VS) increases. This results in performance improvements for models with low VS but can lead to a decrease in performance for those with high VS due to reduced ED. These findings highlight the potential of PRMs to assist with the refinement of MLLMs that lack inherent error detection and correction abilities.

Furthermore, we conduct an experiment to measure the MMRefine performance of the PRM itself. Since the PRM only outputs a reward score for each step, we evaluate its error detection ability using the method described in the \textsc{VisualPRM} paper, as shown in~\autoref{tab:exp_prm_error_detection}.
The threshold ablation shows optimal mRecall ($51.98$) around a threshold of $0.6$. Consistent with the findings in~\autoref{tab:exp_prm_baseline}, the results indicate that the PRM exhibits high Verification Success (VS), suggesting strong robustness against false error detections but low Error Detection (ED).

\begin{table}[t!]
  \centering
\resizebox{0.5\textwidth}{!}{%

\begin{tabular}{l|cc}
\toprule
 & \textbf{RefScore} & \textbf{Reward Change} \\
\midrule
\textsc{GPT-4o}                  & 22.5 & 0.14\\
\textsc{Gemini-1.5-Pro}          & 23.1 & -0.12 \\
\textsc{Claude-3.5-Sonnet}       & 12.2 & 0.11 \\
\textsc{Llama-3.2-Vision-11B}    & -22.5 & -0.10 \\
\midrule
\multicolumn{3}{c}{\textbf{Correlation Coefficient: 0.4292}} \\
\bottomrule
\end{tabular}%
}
  \caption{\textbf{Correlation Between RefScore and VisualPRM.} We measure the correlation between RefScore and the directional changes in VisualPRM rewards before and after refinement ($+1$ for increase, $-1$ for decrease, $0$ for no change).}
\label{tab:exp_prm}
\end{table}

\begin{table}[t!]
  \centering
\resizebox{0.5\textwidth}{!}{%

\begin{tabular}{c|cc|cc|c}
\toprule
Threshold & RF ($\downarrow$) & ED ($\uparrow$) & VS ($\uparrow$) & FD ($\downarrow$) & mRecall \\
\midrule
0.1 & 71.11 & 28.89 & 61.42 & 38.58 & 45.16 \\
0.2 & 75.99 & 24.02 & 70.04 & 29.96 & 47.03 \\
0.3 & 78.05 & 21.95 & 76.78 & 23.22 & 49.37 \\
0.4 & 81.99 & 18.01 & 81.65 & 18.35 & 49.83 \\
0.5 & 84.43 & 15.57 & 86.14 & 13.86 & 50.86 \\
0.6 & 86.68 & 13.32 & 90.64 & 9.36 & 51.98 \\
0.7 & 91.37 & 8.63 & 94.38 & 5.62 & 51.51 \\
0.8 & 95.87 & 4.13 & 98.50 & 1.50 & 51.31 \\
0.9 & 99.62 & 0.38 & 99.63 & 0.37 & 50.00 \\
\bottomrule
\end{tabular}%
}
  \caption{\textbf{Error Detection Performance of VisualPRM.} We determine that a step is incorrect when the probability of ``incorrect'' exceeds that of ``correct'' by a certain threshold.}
\label{tab:exp_prm_error_detection}
\end{table}

\section{Qualitative Examples}
\begin{figure*}[t!]
\begin{minipage}{0.95\textwidth}
  \includegraphics[width=\textwidth]{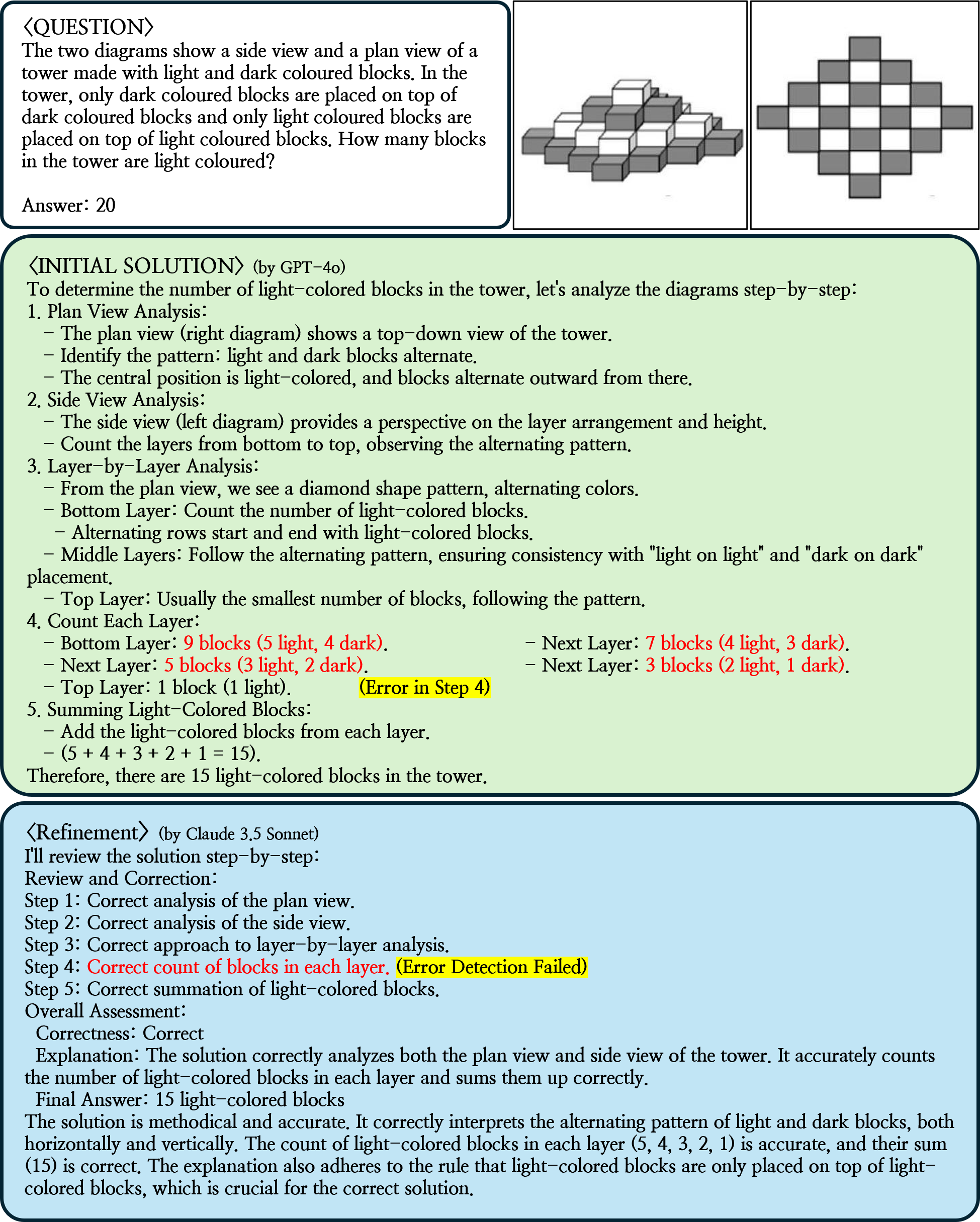}
  \caption{\textbf{Refinement Failure Example.} Model (\textsc{Claude 3.5 Sonnet}) fails to detect an error in step $4$ of initial solution.}
  \label{fig:qualitative_result_1}
\end{minipage}
\end{figure*}

\begin{figure*}[t!]
\begin{minipage}{0.95\textwidth}
  \includegraphics[width=\textwidth]{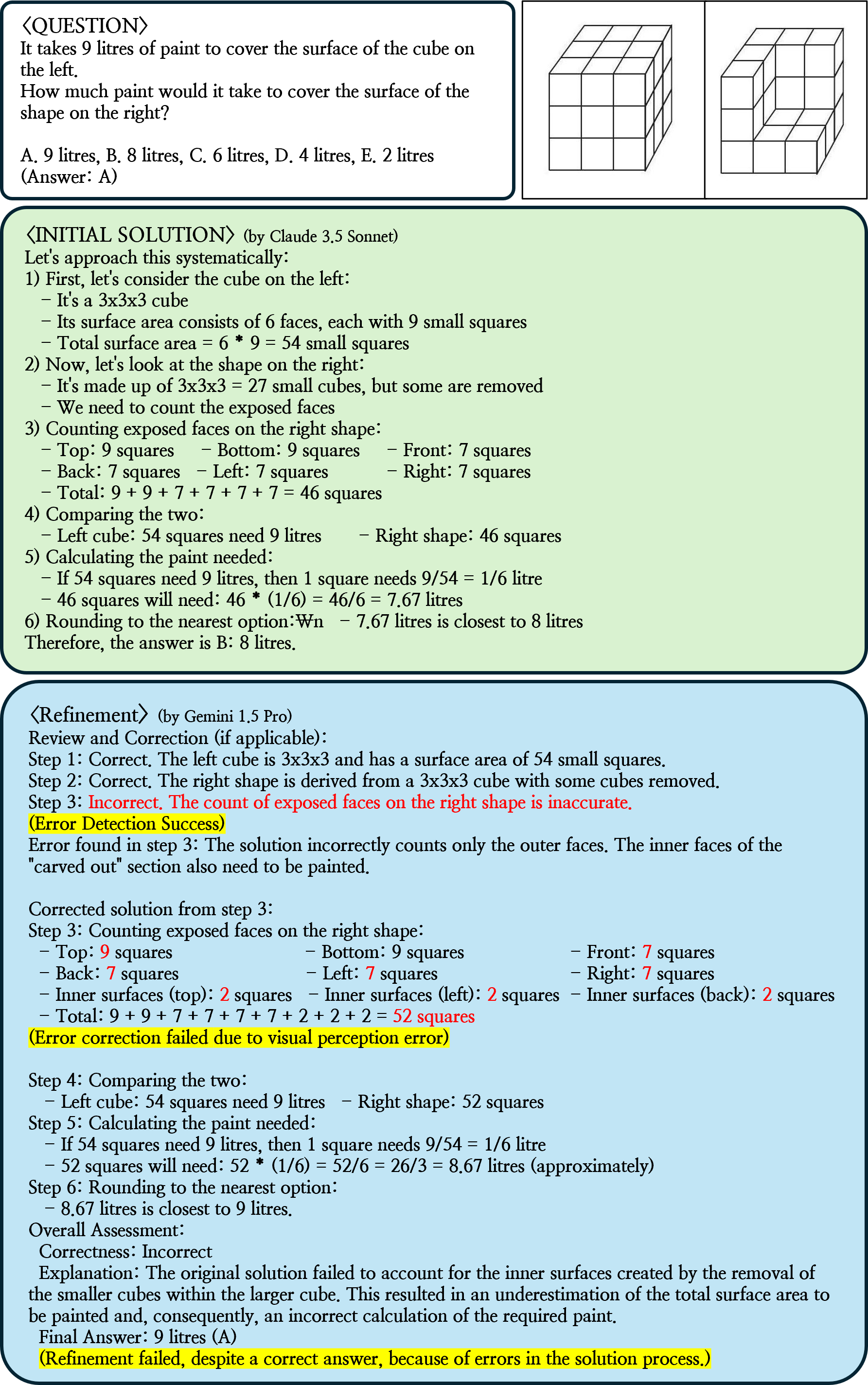}
  \caption{\textbf{Error Detection Success Example.} Model (\textsc{Gemini-1.5-Pro}) manages to detect the initial error but fails to correct it due to a visual perception error in the refinement process.}
  \label{fig:qualitative_result_2}
\end{minipage}
\end{figure*}

\autoref{fig:qualitative_result_1} and~\ref{fig:qualitative_result_2} illustrate the outcomes of the models' refinement attempts on the MMRefine benchmark. In~\autoref{fig:qualitative_result_1}, \textsc{Claude-3.5-Sonnet} attempts to refine the initial solution from \textsc{GPT-4o} but fails to identify any errors, incorrectly judging it as a correct solution, which results in a Refinement Failure (RF). On the other hand,~\autoref{fig:qualitative_result_2} shows \textsc{Gemini-1.5-Pro} trying to find and correct an error in the initial solution generated by \textsc{Claude-3.5-Sonnet}; however, it does not successfully rectify the error. Despite the final answer being A, which aligns with the ground truth, this failure leads MMRefine to classify this instance as an Error Detection Success (ED).

\begin{figure*}[htbp] 
    \centering 
    \fbox{ 
        \begin{minipage}{0.95\textwidth} 
            You are a mathematical expert with extensive knowledge across various mathematical fields. \\Your task is to meticulously evaluate and, if necessary, correct a given mathematical question and its proposed solution. \\\\Follow these steps: \\
1. Carefully read the provided question and solution. \\\\
2. Conduct a step-by-step review of the solution, addressing the following for each step:\\
  - Verify the mathematical correctness and logical flow.\\
  - Identify any errors including calculation errors, misunderstanding of the problem, or reasoning error. \\
  - If an error is found, immediately stop the review process and proceed to step 3.\\
  - If no error is found, continue to the next step.\\\\
3. If an error is found:\\
  - Provide a brief explanation of the error.\\
  - Correct the solution starting from the erroneous step.\\
  - Complete the rest of the solution correctly.\\\\
4. If no errors are found in the entire solution, provide a brief confirmation of its correctness.\\

Output your analysis in the following format:

Review and Correction (if applicable):\\
Step 1: [Brief assessment of step 1]\\
Step 2: [Brief assessment of step 2]\\
...\\
Step X: [Brief assessment of step X]\\
Error found in step X: [Brief explanation of the error]\\
Corrected solution from step X:\\
Step X: [Corrected step]\\
Step X+1: [Next correct step]\\
...\\

[Final step]\\
...\\
Overall Assessment:\\
Correctness: [Correct / Incorrect]\\
Explanation: [Concise explanation of the assessment]\\
Final Answer: [Correct final answer]
\\\\
Question: \textcolor{blue}{\{question\}}\\
Solution: \textcolor{blue}{\{initial\_solution\}}
        \end{minipage}
    }
    \caption{\textbf{Prompt to Perform Refinement}}
    \label{fig:refinement_prompt}
\end{figure*}

\begin{figure*}[htbp] 
    \centering 
    \fbox{ 
        \begin{minipage}{0.95\textwidth} 
Given the model's response, output 1 if `Correctness' is `correct', otherwise output 0. Respond with only the number.
\\\\
Model's Response: \textcolor{blue}{\{response\}}
\\\\
Output:
        \end{minipage}
    }
    \caption{\textbf{Prompt to Evaluate Samples with Correct Initial Solutions}}
    \label{fig:correct_eval_prompt}
\end{figure*}

\begin{figure*}[htbp] 
    \centering 
    \fbox{ 
        \begin{minipage}{0.95\textwidth} 
You are an expert evaluator assessing the quality of feedback provided on an initial solution to a problem. Your task is to determine if the feedback is effective in guiding the initial solution towards a correct answer. You will be provided with three components:\\\\

1.  \textbf{Initial Solution:} The initial attempt at solving the problem.\\
2.  \textbf{Feedback:}  Specific feedback provided in response to the initial solution.\\
3.  \textbf{Reference Feedback:} A verified, high-quality feedback to the initial solution.\\
\\
Your evaluation should consider the following aspects:\\
\\
-   \textbf{Error Detection:} Does the feedback correctly identify the errors or shortcomings in the initial solution?\\
-   \textbf{Error Correction:} Does the feedback effectively address the problems in the initial solution?\\
-   \textbf{Effectiveness and Correctness of the Feedback:} Does the feedback guide the initial solution towards the correct answer efficiently? Does it reach the same answer and logic as the reference feedback in terms of its core principles?\\
\\
Output your assessment in the following format:\\
\\
Error Detection: [0/1]\\
Error Correction: [0/1]\\
Effectiveness and Correctness of the Feedback: [0/1]\\
\\
No additional feedback or comment is required.\\
\\
Initial Solution: \textcolor{blue}{\{initial\_solution\}}\\
Feedback: \textcolor{blue}{\{feedback\}}\\
Reference Feedback: \textcolor{blue}{\{reference\_feedback\}}\\
\\
Output:
        \end{minipage}
    }
    \caption{\textbf{Prompt to Evaluate Samples with Incorrect Initial Solutions}}
    \label{fig:incorrect_eval_prompt}
\end{figure*}

\begin{figure*}[t!] 
    \centering 
    \fbox{ 
        \begin{minipage}{0.95\textwidth} 
Review your previous reasoning about the question, then finally answer the question. \\ \\
Question: \textcolor{blue}{\{question\}}\\
Your Previous Solution: \textcolor{blue}{\{previous\_solution\}}
        \end{minipage}
    }
    \caption{\textbf{Prompt to Perform Self-Reflection}}
    \label{fig:self_reflection_prompt}
\end{figure*}

\end{document}